\documentclass{svproc}
\usepackage{url}

\usepackage{graphicx}
\graphicspath{{./images/}}
\usepackage{acronym}
\usepackage{csquotes}
\usepackage{hyperref}
\usepackage{siunitx}
\usepackage{subcaption}
\usepackage{todonotes}

\usepackage{amsmath}
\usepackage{amsfonts}
\usepackage{amssymb}
\usepackage{float}

\begin{document}
\mainmatter              
\title{Model-agnostic Body Part Relevance Assessment for Pedestrian Detection}
\titlerunning{Body Part Relevance Assessment}  
%
\author{
    Maurice Günder\inst{1,2} \and 
    Sneha Banerjee\inst{1,3} \and 
    Rafet Sifa\inst{1,2} \and 
    Christian Bauckhage\inst{1,2} 
}
\authorrunning{M. Günder et al.} 
%
%
\institute{
    Fraunhofer Institute for Intelligent Analysis and Information Systems IAIS, Schloss Birlinghoven, 53757 Sankt Augustin, Germany \\
    \email{maurice.guender@iais.fraunhofer.de}
    \and
    University of Bonn, Institute for Computer Science III, Friedrich-Hirzebruch-Allee 5, 53115 Bonn, Germany
    \and
    RWTH Aachen University, Department of Computer Science, Ahornstraße 55, 52074 Aachen, Germany
}
\maketitle              
\begin{abstract}

Model-agnostic explanation methods for deep learning models are flexible regarding usability and availability. However, due to the fact that they can only manipulate input to see changes in output, they suffer from weak performance when used with complex model architectures. For models with large inputs as, for instance, in \acl{OD}, sampling-based methods like KernelSHAP are inefficient due to many computation-heavy forward passes through the model. In this work, we present a framework for using sampling-based explanation models in a computer vision context by body part relevance assessment for pedestrian detection. Furthermore, we introduce a novel sampling-based method similar to KernelSHAP that shows more robustness for lower sampling sizes and, thus, is more efficient for explainability analyses on large-scale datasets. 


\end{abstract}

\section{Introduction}\label{sec:intro}

Today's deep learning model architectures are more powerful than ever and enable the use of \ac{AI} in a wide range of application areas. However, with increasing model complexity comes increasing opacity and their output is less (human-)interpretable. Therefore, it is not uncommon for large models to be regarded only as black boxes. This can be particularly problematic in safety-relevant applications as, for instance, in \ac{AD} where \ac{AI} models cause decisions of autonomous systems that should be trustworthy, reasonable, and explainable~\cite{safety_review}. Thus, the field of \ac{XAI}~\cite{xai_review} is of increasing interest.

Generally, \ac{XAI} approaches for analysis of deep learning models can be categorized in \textit{model-specific} and \textit{model-agnostic} methods. While model-specific methods are tailored to the underlying architecture and manipulate the test model in inference and/or training, model-agnostic methods are applied in a post-hoc manner to the test model, i.e., to fully trained models. These methods have the advantage of high flexibility, since models are treated as black boxes and, thus, any model can be analyzed the same way. Hence, the interpretation or explanation results can be compared across model classes or architectures. However, a major drawback of model-agnostic \ac{XAI} methods is that only the model input can be manipulated to analyze consequential output changes. Therefore, these methods are sampling-based, which leads to a high computational effort for complex models.

In \ac{AD}, the trustworthy recognition of street scenes, especially pedestrians~\cite{ped_det_review1,ped_det_review2}, is of major interest. Contemporary \ac{OD} models show good performances, but have very distinct basic architectures and working principles.~\cite{od_survey} For pedestrian detection, a severe challenge is that commonly, pedestrians appear under occlusion so that \ac{OD} models have high robustness requirements here.~\cite{pd_under_occlusion} From an \ac{XAI} point of view, it is therefore of particular importance on which semantic regions a test model bases its decisions for detecting a pedestrian, regardless of the underlying test model. Hence, model-agnostic explanation models should be considered here since it enables high flexibility and comparability. Many semantic knowledge concepts that appear on pedestrians like clothing, accessories, or poses mostly coincide with specific body parts and the division into body parts is coherent among different perspectives. Only the visibility of individual body parts differs between individual pedestrian instances. Therefore, we can use body parts as semantic regions in order to profile and benchmark object detection models with each other.

Model-agnostic explanation methods can be further distinguished into global and local explanation methods. Global methods try to explain the model on the data as a whole to interpret the overall performance, whereas local methods try to explain outputs for single data points or instances. \ac{TCAV} is a method that tests a model for relevant features that are given by example images~\cite{tcav}. A \ac{CAV} then quantifies the extent to which the model was activated in a prediction to a given concept. 
Those concepts can be, for instance, textures, color schemes, or anything that is describable by a bunch of images. Since we want to assess body parts, we do not have clear color or textures we want to focus on, and example images of isolated body parts are not available. This is why we do not focus on the \ac{TCAV} in this work. Rather, we will focus on a formalism called \ac{LIME}~\cite{lime}. \ac{LIME} is a method for local explanation of instances by introducing a surrogate model that is simpler and more interpretable than the typically complex reference model~\cite{lime}. Further prerequisites established a formalism called \ac{SHAP} presented by Lundberg and Lee~\cite{shap}. The approach in this work is based on Kernel\ac{SHAP}, a specification of \ac{SHAP} that enables local model-agnostic explanations, so that we go into further details in Section~\ref{sec:materials_and_methods}. 

However, when it comes to \ac{OD} problems, model-agnostic explanation methods show some shortcomings being based on input sampling. In comparison to many machine learning tasks dealing with image processing, the input dimension and typically the model size is rather low, which makes single forward passes through the model quite fast. In image processing or, particularly, \ac{OD} tasks, the input, i.e., image data, is rather complex and forward passes are computationally heavier. Thus, sampling images causes lots of forward passes decelerating the model explanation substantially. Due to the drastically larger number of input dimension, even more samples are needed to gain meaningful model explanations.

Therefore, we need to adapt sampling-based, model-agnostic explanation methods like Kernel\ac{SHAP} to explain the output pedestrian detection models.

\section{Related Works in \ac{XAI}}

In recent years, the domain of \ac{XAI} has gathered a significant momentum, particularly in the field of image processing. This surge is driven by the critical need to enhance transparency, accountability, and trust in \ac{AI} systems, especially those deployed in sensitive domains like healthcare, autonomous vehicles, and security. Here, we review some prominent works in the realm of \ac{XAI} in image processing. The existing works in the literature can be broadly categorized into five main fields based on the type of models that has been used for the purpose: interpretable \ac{CNN}s, attention mechanism based models, decision trees and rule-based models, \ac{GAN}s for explainability, \ac{CBR} and prototypical networks.

\paragraph{Interpretable \ac{CNN}s \textemdash}
Zeiler and Fergus introduced the concept of \enquote{deconvolution networks} in~\cite{zeiler2014visualizing}, enabling visualization of feature activations to elucidate \ac{CNN} decisions. Zhang et al. in~\cite{zhang2018interpretable} proposed the concept of an interpretable \ac{CNN} for better understanding of the representations of the higher convolution layers in a \ac{CNN}. A special loss for each of the filters in the higher convolution layers were used so that each of these filters of an interpretable \ac{CNN} corresponds to a distinct part of the object. It mitigated the need for manual object part annotations, which are often unavailable in real datasets. Selvaraju et al. proposed \ac{Grad-CAM} \cite{selvaraju2016grad}, facilitating the localization of discriminative regions in images influencing \ac{CNN} predictions. By displaying the input regions that are \enquote{important} for predictions by heatmaps, it increased the transparency of \ac{CNN}-based models by providing visual explanations. But the application of the previously mentioned methods were limited to visual explanations and could not capture more complex decision making factors which were not specific to a particular region of an object (e.g., properties of the scene like weather of the outdoor scene)~\cite{rudin2022interpretable}.

\paragraph{Attention mechanism \textemdash}
Xu et al. pioneered the application of attention mechanisms in image captioning in~\cite{xu2015show}, allowing networks to focus on salient image regions during prediction. Fukui et al. extended this idea with \ac{ABN}~\cite{fukui2019attention}, augmenting \ac{CNN}s with attention modules to enhance interpretability. In the previously mentioned works, the learned weights were used to display the attended regions of an image or text which was used to verify the mechanism they were designed to employ. Kim et al. show in~\cite{kim2017visual} that Hadamard product in multimodal deep networks implicitly carries on an attention mechanism for visual inputs. They demonstrate how the Hadamard product in multimodal deep networks takes into account both visual and textual inputs simultaneously by using a gradient-based visualization technique and has a superior performance as compared to the respective learned attention weights~\cite{rodis2023multimodal}. But using attention mechanism for images can be often computationally expensive, thus having a limitation on its scalability~\cite{hou2021coordinate}. 

\paragraph{Decision Trees and rule-based models \textemdash}
Zhang et al. introduced decision tree guided \ac{CNN}s in~\cite{Zhang_2019_CVPR}, integrating decision trees with \ac{CNN}s to provide explicit reasoning for classification scores. They suggested learning a decision tree which provided a semantic explanation for each prediction given by the \ac{CNN}. The feature representations in higher convolution layers are broken down into fundamental concepts of the object parts by the decision tree. In this manner, it indicates which part of the object activate which prediction filters, as well as the relative contribution of each object part to the prediction score. The decision tree explains \ac{CNN} predictions at various fine-grained levels by arranging all possible choice made in a coarse-to-fine order. These semantic justifications for \ac{CNN} predictions have enhanced importance that is not just limited to the conventional pixel-level analysis of \ac{CNN}s. But such an explanation method is dependent on the model. Though some model-specific explanation techniques may be more helpful in certain situations for a given model than model-agnostic techniques, but the latter is more scalable. It has the benefit of being totally independent to the model, retaining the ability to apply these techniques in whole different use cases where the predictive model is different~\cite{carvalho2019machine}. Also, decision trees might lack the expressive power required to capture complex patterns in high dimensional data like images. Linear regression models are a type of rule-based models that defines a relationship between the inputs and the outputs of the model by fitting a linear equation to the observed data. The \ac{LIME}~\cite{lime} method, that is mentioned already in Section~\ref{sec:intro}, uses surrogate linear models for the explainability of the black-box model.

\paragraph{\ac{GAN}s for explainability \textemdash}
The majority of interpretable \ac{GAN} applications~\cite{bau2018gan,shen2020interfacegan,chen2016infogan} at this time deal with creating and altering images. These applications are limited by the kinds of datasets and resources that can be used to train GAN, as well as the application scenarios and techniques that are needed for specific tasks. Interpretable techniques therefore have a limited degree of generalization. Certain high-risk domains, like software and intrusion detection, malicious speech, disease diagnosis, and mortality predictions, involve less \ac{GAN} application. There is also a lack of a cohesive and consistent interpretable framework~\cite{genovese2019towards,li2020interpreting} in the research of \ac{GAN} interpretability, and it is heavily dependent on particular problems, task scenarios, and models, leading to a low level of universality for interpretable approaches. Increasing model transparency is necessary to investigate the interpretability of GAN models. Privacy protection is at risk if the data is transparent. Therefore, a major issue for the current GAN interpretable approaches is to improve the interpretation effect in certain high risk applications like healthcare, security, autonomous driving while maintaining the security of GAN models and data privacy.~\cite{wang2023current}

\paragraph{\ac{CBR} and prototypical networks \textemdash}
Prototypical Networks have several important advantages. What distinguishes them is their capacity to generalize well from small amounts of labelled input, which makes them especially useful in few-shot learning contexts~\cite{snell2017prototypical}. In situations where data scarcity presents a difficulty, Prototypical Networks perform better than standard models, which call for large amount of labeled data. Furthermore, prototype-based method supports strong classification, improving generalization across different domains. These networks' ability to quickly adjust to new data is another benefit of the iterative learning mechanism, which strengthens their standing as flexible and adaptive machine learning models. Prototypical networks do have certain drawbacks, though. The use of embedding space and distance metrics, which cannot always adequately portray the intricate relationships between data points, is one main area of concern. The representativeness and quality of the labeled training data can also have an impact on how effective these networks are. Furthermore, Prototypical Networks perform best in few-shot scenarios~\cite{snell2017prototypical}, but may struggle to define prototypes in use-cases with a high degree of complexity or diversity of classes~\cite{davoodi2023interpretability}. \ac{CBR}-\ac{CNN} combines \ac{CNN}s for feature extraction with \ac{CBR} for decision-making in image classification. It begins by extracting features from input images using a \ac{CNN}, then retrieves similar cases from a database based on these features~\cite{khan2019hybrid}. The final classification decision is made by aggregating the classifications of retrieved cases. This approach leverages both the deep learning capabilities of \ac{CNN}s and the knowledge-driven reasoning of \ac{CBR} to enhance the interpretability and performance of image classification systems~\cite{louati2021deep}. But, these approaches rely on a diverse and representative case database for generalization, increased computational complexity due to hybrid architecture, and sensitivity to retrieval mechanisms~\cite{safa2022survey}. Capturing complex semantic relationships, dependency on annotated data, and adaptability to dynamic environments pose additional challenges~\cite{keane2019case}.

\section{Materials and Methods}\label{sec:materials_and_methods}

\begin{figure}[tbh]
    \centering
    \includegraphics[width=\textwidth]{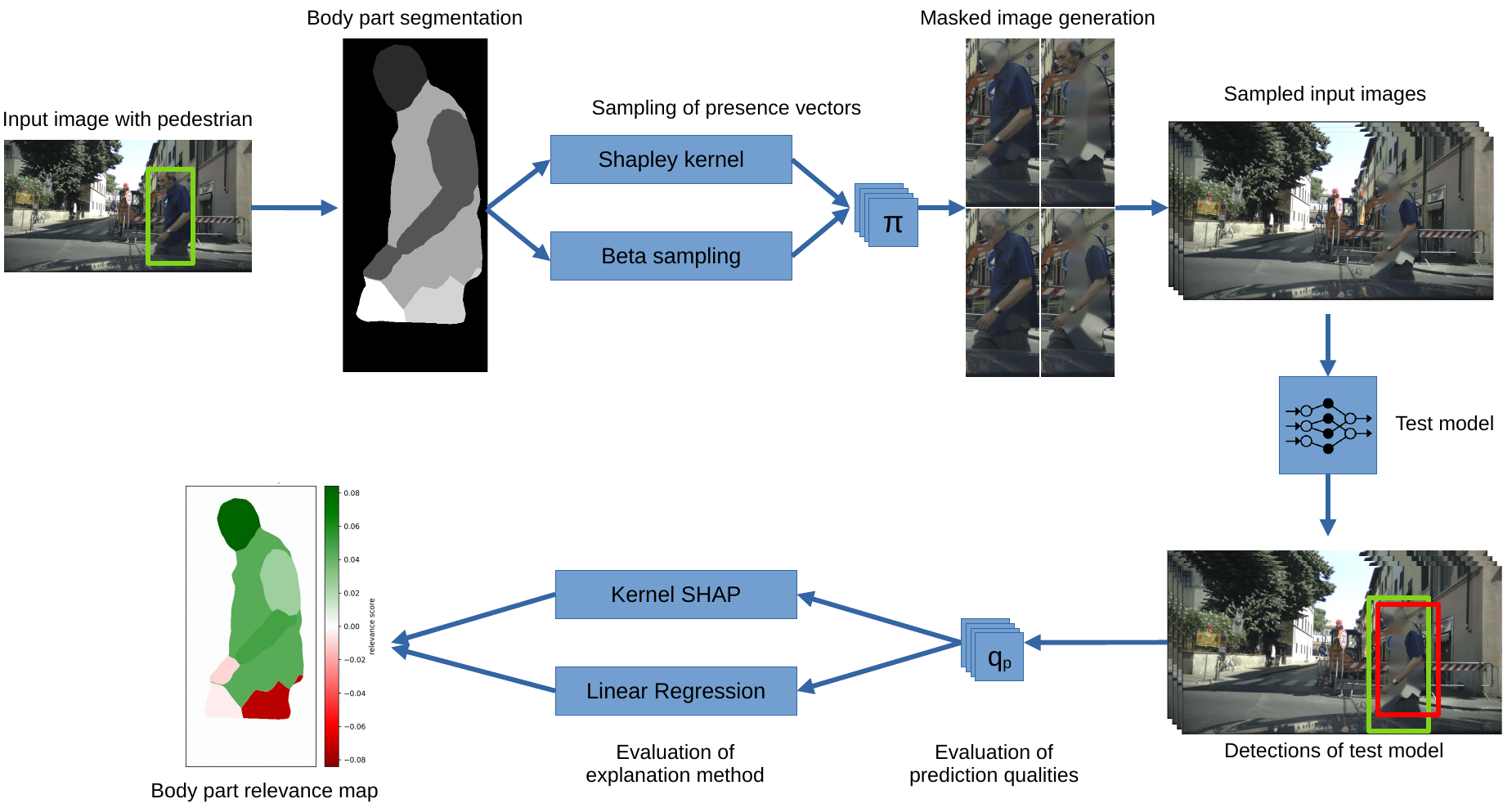}
    \caption{Concept overview of our approach to model-agnostic body part relevance assessment.}
    \label{fig:concept_sketch}
\end{figure}

We now shed light on how our approach to model-agnostic body part relevance assessment is structured. Figure~\ref{fig:concept_sketch} outlines the concept from the input street scene image to the so-called relevance maps. The details about the individual modules shown in this sketch are explained in the next sections.

\subsection{\acf{SHAP}}

We already mentioned the \ac{LIME} method briefly in Section~\ref{sec:intro}. In this paragraph, we want to go into further details in order to shed light on the explanation procedure and how it is connected to \ac{SHAP}. 

Mathematically spoken, LIME generally minimizes the objective function
\begin{equation}\label{eq:lime}
    \xi = \underset{g\in\mathcal{G}}{\arg\min{}} L(f,g,\pi_{x'}) + \Omega(g)\,
\end{equation}
where $L$ is the loss over a sample set in the interpretable space given $f$ as the original model, $g$ as the local explanation model in a set $\mathcal{G}$ of possible models, and $\pi_{x'}$ as a proximity measure, or kernel, between local instances.
The surrogate model $g$ can be chosen arbitrarily, which allows a lot of freedom in modeling but could end up in a surrogate model that is not human-interpretable. Therefore, a penalty $\Omega(g)$ is added to the loss function to avoid unnecessary complex surrogates.

With their approach called \ac{SHAP}, Lundberg and Lee~\cite{shap} proposed a set of method that should unify explanation approaches by incorporating properties of so-called Shapley values~\cite{shapley,game_theory_regression} into \ac{LIME}. Shapley values originate from game theory and are well-defined and theoretically based measures for feature contribution to a certain outcome. The formalism of Shapley values can be incorporated in a linear \ac{LIME} model by a so-called \enquote{Shapley kernel} to calculate the contribution of each feature to the output. This approach is called Kernel\ac{SHAP}. Skipping over some particulars that are present in the initial work~\cite{shap}, the Shapley kernel sets the terms of equation~\eqref{eq:lime} as
\begin{align}
    \Omega(g) &= 0\,\label{eq:shapley_kernel:regularization}\\
    \pi_{x'}(z') &= \frac{M-1}{{M\choose |z'|} |z'| (M-|z'|)}\,\label{eq:shapley_kernel:weighting}\\
    L(f,g,\pi_{x'}) &= \sum\limits_{z'\in Z}\left[ f(h_x^{-1}(z')) - g(z') \right]^2 \pi_{x'}(z')\,\label{eq:shapley_kernel:loss}
\end{align}
where $h_x$ is a mapping between the complex and the explanation model, i.e., it is $g(x') = f(x)$ when $x = h_x(x')$. The $M$ input features $z'\in {0,1}^M$ are binary (\enquote{present} or \enquote{absent}), so that $|z'|$ denotes the number of present features. Furthermore, the Kernel\ac{SHAP} method preserves that the Shapley values can be solved by linear regression without special restrictions on the original model. Thus, it is model-agnostic.

\subsection{Superpixel Model}

In image processing like \ac{OD}, the input size is typically much larger than for other machine learning tasks. Those large input sizes make sample based analyses mostly infeasible due to the large combinatorial space. This is why the input size has to be drastically reduced in order to have efficient sampling. Moreover, the contribution of a single pixel to the actual detection can be considered to be negligibly small. Thus, a commonly used trick is to summarize a region of image pixels as a so-called \textit{superpixel}. One way would be a fixed tiling into rectangular or quadratic superpixels, ignoring the actual image content. The other way is to define superpixels by semantic regions with similar texture, color, shape, or, in our case of pedestrian detection, body parts. In contrast to the fixed tiling, the semantic regions usually have different sizes.

The superpixels now serve as the mapping $h_x$ between the large pixel and the smaller superpixel input space. Based on that, the Kernel\ac{SHAP} method estimates the attribution of the input features to the output. Thus, we need to parametrize the superpixels by feature values. Our superpixel model, which serves as the explainable surrogate model, should have interpretable feature values. As we want to assess the relevance of body parts to the pedestrian detection, the feature values should represent the degree of information that is visible in the respective superpixel. Therefore, we introduce a \textit{presence} value $\pi_i$ for each superpixel $i$. A value of $\pi_i = \num{1}$ means that the $i$-th superpixel is fully visible, as in the original input image. With decreasing presence value $\pi_i \to \num{0}$, the superpixel gets increasingly hidden. In this work, we use three methods to hide the information of the superpixel. The first method is to overlay the superpixel with noise sampled from the information of the remaining image by a multinomial normal distribution given by the RGB information. The second method overlays the superpixel with noise sampled from the information of the neighboring superpixel contents. The third method is to remove all superpixels by a content-away \textit{inpaint} method implemented in the OpenCV library~\cite{opencv}. A presence value of $\pi_i = \num{0}$ means, that only the overlay is visible in the image, i.e., the superpixel information is completely hidden. 

\begin{figure}[tbh]
    \centering
    \includegraphics[width=0.8\textwidth]{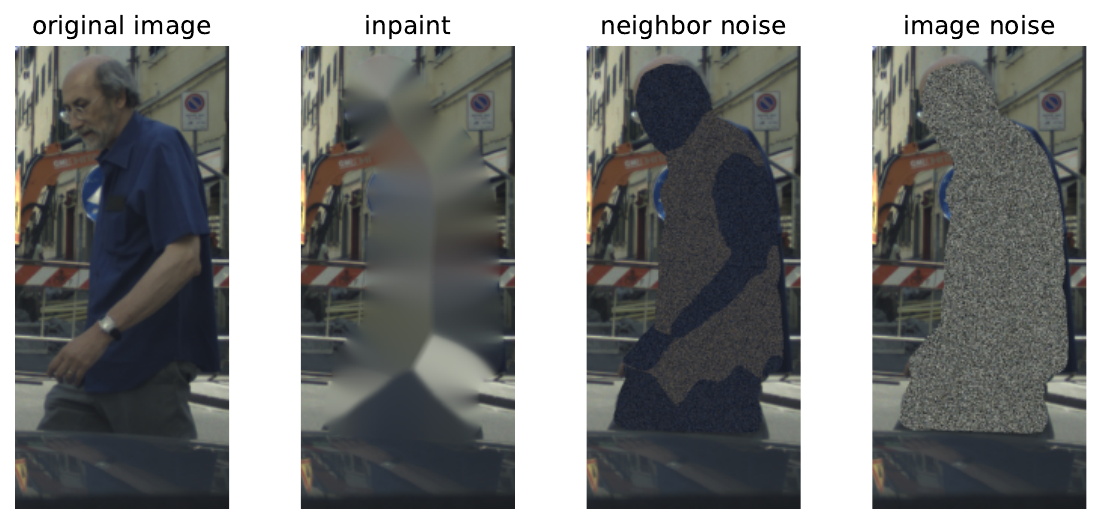}
    \caption{Comparison of our masking methods demonstrated on a pedestrian image from the EuroCity Persons dataset~\cite{eurocitypersons}.}
    \label{fig:mask_methods}
\end{figure}

Thus, our superpixel model for the body part relevance assessment gets a presence vector $\vec{\pi}\in[0,1]^k$ for $k$ visible body parts as an input and samples an image based on this vector. This image is forwarded to the black-box \ac{OD} model that should be analyzed. Figure~\ref{fig:mask_methods} shows our three masking methods in the case of fully hidden body parts, i.e., $\vec{\pi} = \vec{0}$.

The typical output of an \ac{OD} model are labels, \ac{bbox} coordinates, and classification scores. The number of those elements is dependent from the number of detected objects in the input image. Thus, we need to formalize the detection quality of a distinct pedestrian of interest among multiple possible detections with multiple \acp{bbox} and scores. For pure classification, the classification score would be enough, but for \ac{OD}, it is desirable for a detection quality score to include information about the precision of the \ac{bbox}, as well. Therefore, we calculate the \ac{DICE} between our ground truth \ac{bbox} and all detection's \acp{bbox} defined by
\begin{align}
    \ac{DICE}(A,B) = \frac{2|A\cap B|}{|A|+|B|}\,,
\end{align}
where $A$ and $B$ are the two \acp{bbox} of interest. We identify the correct \ac{bbox} by the maximum \ac{DICE} with the ground truth \ac{bbox} $G$. To include also the pure classification quality, we multiply this value with the respective classification score $c$ for the detection. Thus, our detection quality $q_p$ of a pedestrian $p$ with detected bounding box $P$ is
\begin{align}
    q_p = \ac{DICE}(P,G) \cdot c_p\,.
\end{align}
Since \ac{DICE} and $c_p$ are values in the interval $[0,1]$, it is $q_p \in [0,1]$.

All in all, we now wrapped our \ac{OD} model into a surrogate superpixel model, with an input vector and an output scalar.

\subsection{Body Part Segmentation}

\begin{figure}[tbh]
    \centering
    \begin{subfigure}[t]{0.235\textwidth}
        \centering
        \includegraphics[width=0.35\textwidth]{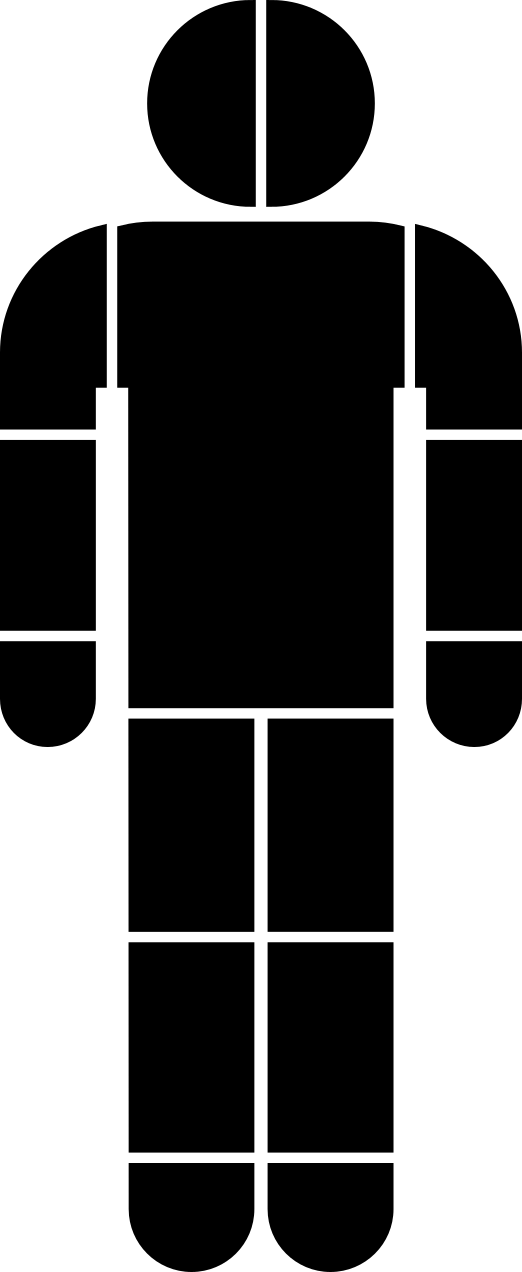}
        \subcaption{Level 0. This is the original output of the BodyPix model. It has in total 24 body parts including front and back for the arm, leg, and torso parts. The orientations are w.r.t. the ego perspective.}
        \label{fig:abst_levels:0}
    \end{subfigure}
    \hfill
    \begin{subfigure}[t]{0.235\textwidth}
        \centering
        \includegraphics[width=0.35\textwidth]{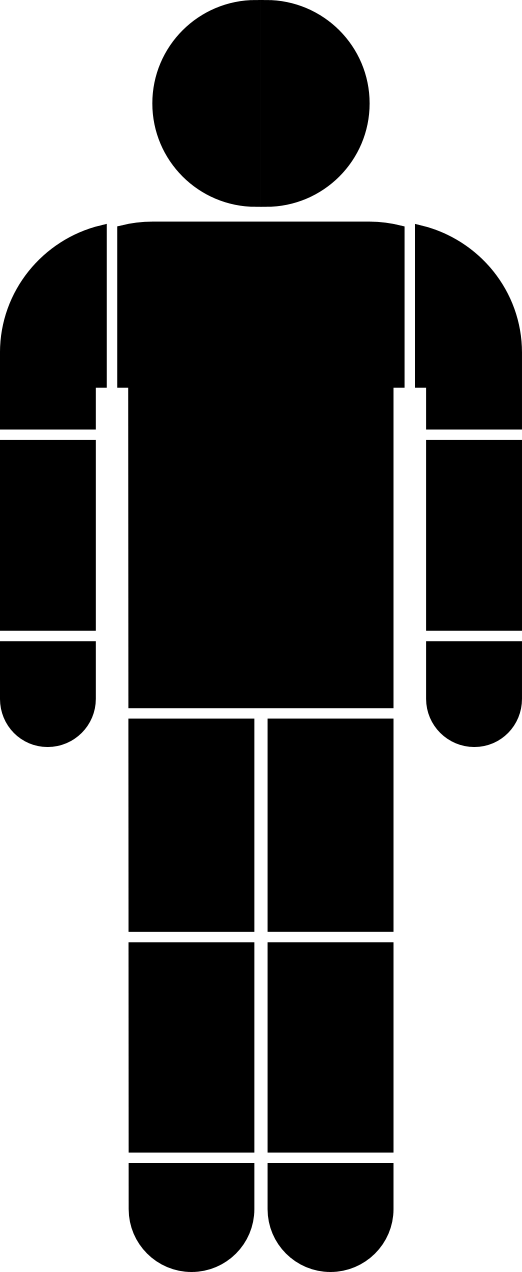}
        \subcaption{Level 1. In tiis first abstraction level, the two face halves are unified. Additionally, there is no differentiation of front and back parts any more. Overall, this results in 14 body parts.}
        \label{fig:abst_levels:1}
    \end{subfigure}
    \hfill
    \begin{subfigure}[t]{0.235\textwidth}
        \centering
        \includegraphics[width=0.35\textwidth]{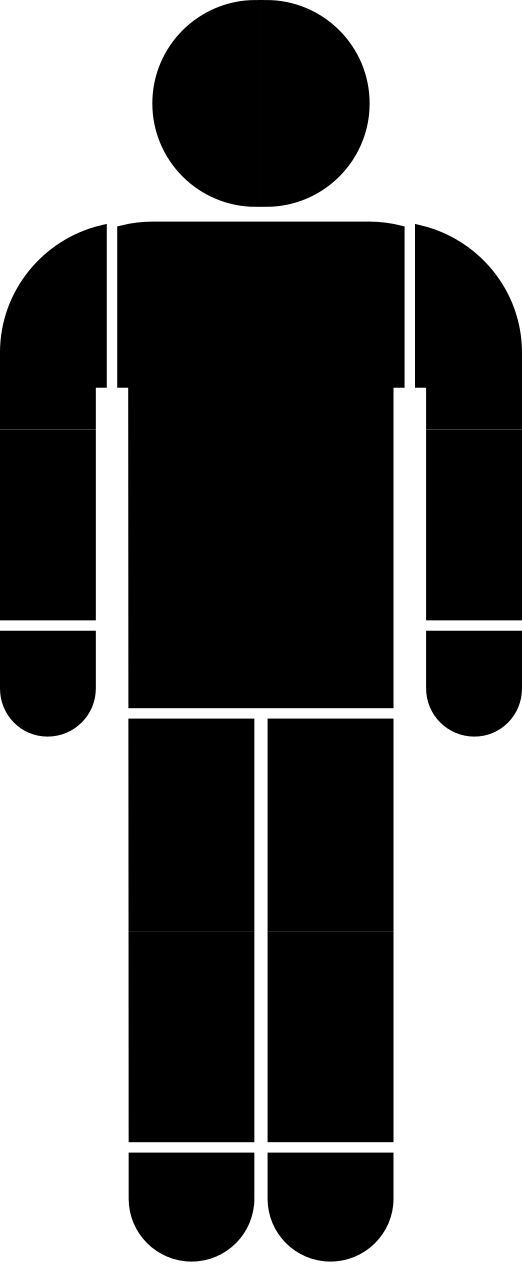}
        \subcaption{Level 2. In this second abstraction level, the upper and lower parts of arms and legs are unified, as well, resulting in 10 remaining body parts.}
        \label{fig:abst_levels:2}
    \end{subfigure}
    \hfill
    \begin{subfigure}[t]{0.235\textwidth}
        \centering
        \includegraphics[width=0.35\textwidth]{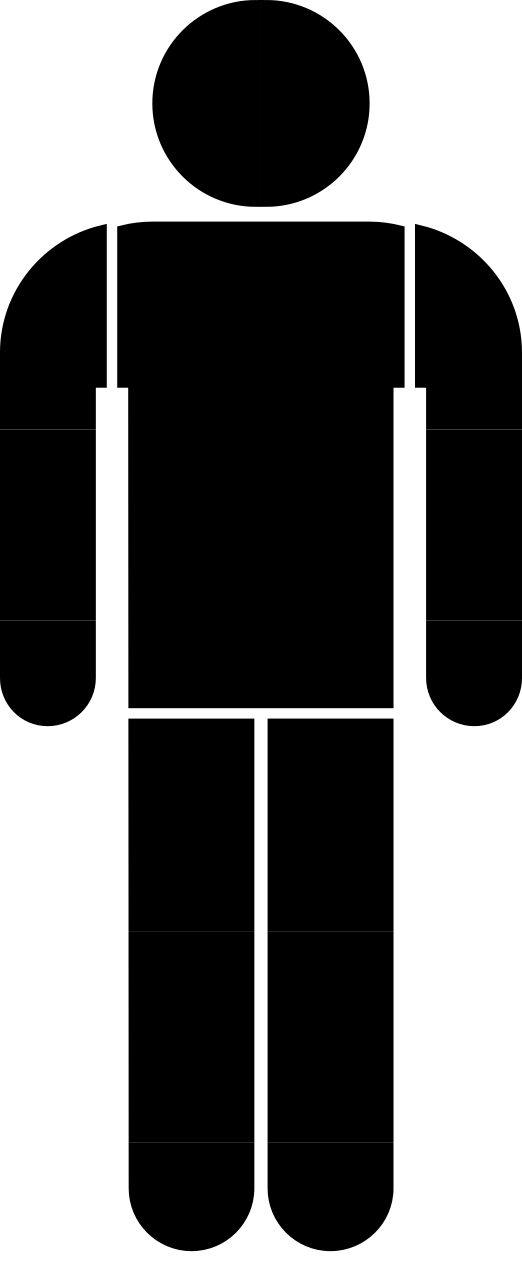}
        \subcaption{Level 3. In this third abstraction level, hands are unified with the arms and feet are unified with the legs resulting in 6 remaining body parts.}
        \label{fig:abst_levels:3}
    \end{subfigure}
    \caption{Abstraction levels of our body part segmentation. The levels represent the granularity from detailed (\subref{fig:abst_levels:0}) to less detailed (\subref{fig:abst_levels:3}).}
    \label{fig:abst_levels}
\end{figure}

In order to introduce the superpixel model parametrization to our pedestrian detection model, we need to get a segmentation of the body parts. For the currently available large-scale pedestrian datasets like CityPersons~\cite{citypersons} or EuroCity Persons~\cite{eurocitypersons}, proper body part segmentations are not available. Thus, we utilize \textit{BodyPix}, a trained model for body segmentation~\cite{bodypix_github}. BodyPix enables us to have vast amounts of real world pedestrian data. However, two major drawbacks are that the segmentation quality is rather low if the pedestrians resolution is low, i.e., for pedestrians appearing far away in the image. The other major drawback is that there is no instance segmentation available, which means that for pedestrian groups or multiple pedestrians in one \ac{bbox}, we can only access the same body parts of all pedestrians at one time. At least, we can reduce the impact of the resolution problem by focusing our relevance assessment only on the biggest pedestrians, measured by \ac{bbox} area, in the dataset of interest. By default, BodyPix segments \num{24} different body parts, including front and back parts. We can simplify our analysis by introducing \num{3} further mappings, where body parts are unified. We call those mappings \textit{abstraction levels}, where level \num{0} is the original BodyPix output. The granularity reduces with ascending level number. The mappings are shown in Figure~\ref{fig:abst_levels}.

\subsection{From Sampling to Local Explanation}

In Kernel\ac{SHAP}, one first defines an input and a baseline. The input is the instance to explain, so, in our case, the visible pedestrian, i.e., we set $\vec{\pi}_\text{input} = \vec{1}$ as the input. As the baseline, we set a completely absent or hidden pedestrian, thus it is $\vec{\pi}_\text{baseline} = \vec{0}$. The sampling of the binary perturbation is weighted with the Shapley kernel and feature attribution values are calculated using weighted linear regression.~\cite{shap} In this work, we will call those attribution values \textit{(body part) relevance scores}.

\begin{figure}[tbh]
    \centering
    \includegraphics[width=0.6\textwidth]{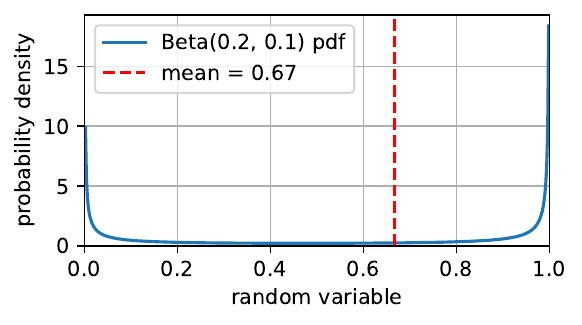}
    \caption{Plot of the Beta distribution (Equation~\eqref{eq:beta_dist}), the presence vectors of our sampling method are drawn from. The red dashed line shows the expectation value (mean) of the distribution.}
    \label{fig:beta_dist}
\end{figure}

As mentioned, Kernel\ac{SHAP} perturbs the instance by masking features, so that all body parts can be absent or present and, hence, it does not consider our still possible partly presences with $0 < \pi_i < 1$. This is due to the Shapley-conform weighting kernel definition in Equation~\eqref{eq:shapley_kernel:weighting} that only considers binary values. Therefore, we introduce a second custom sampling and explanation method using continuous sampling and, as well as Kernel\ac{SHAP}, linear regression to get the scores, but without following the Shapley properties. A uniform sampling of the presence values would end up in many \enquote{blended} body parts which is rather unrealistic. This is why we use a distribution that concentrates on values near \num{0} and \num{1}. One distribution that has this property is the Beta distribution
\begin{align}\label{eq:beta_dist}
    B(x, \alpha, \beta) = \frac{\Gamma(\alpha+\beta)}{\Gamma(\alpha)\Gamma(\beta)} x^{\alpha-1}(1-x)^{\beta-1}
\end{align}
with the so-called concentration coefficients $\alpha$ and $\beta$. By deliberately choosing proper values for $\alpha$ and $\beta$, we can not only steer the concentration strength to the boundaries, but also the expectation value. Without loss of generality, we choose $\alpha = 0.2$ and $\beta = 0.1$ resulting in a distribution that is concentrated on its limits at $0$ and $1$ with a slightly stronger concentration on $1$. The expectation value results in an average pedestrian visibility of about $\SI{67}{\percent}$. An expectation value above $\SI{50}{\percent}$ makes sense in our use case since otherwise, the pedestrian might often be not detected at all, resulting in a detection quality of $\num{0}$. Thus, if too many generated samples end up in non-recognitions, we will not get insights in the relevance of body parts and the sampling becomes inefficient. Figure~\ref{fig:beta_dist} shows a plot of the presence vector sampling \ac{pdf} of our custom method. Another reason to concentrate the \ac{pdf} on the limits is to have a robust linear regression even with a low amount of samples due to many data points at the outermost regions of the regression domain. This is also why too high visibility expectation values are counterproductive as well, since the model will probably detect all sampled instances, and we have fewer counterexamples to get insights into the prediction boundaries of the test model.

Once the presence vectors are sampled and propagated through the superpixel and the \ac{OD} model, we have the corresponding pedestrian detection quality scores and can calculate our body part relevance scores for both explanation methods by linear regression. The relevance scores can be visualized by the body part shapes with colors representing the respective relevance scores. We call those visualization \textit{relevance maps}. Furthermore, we estimate the error of the relevance scores of our method by performing 4 independent regressions with a subset of \SI{75}{\percent} of all data points. For each regression, we draw a different random subset. This method is commonly called \enquote{bootstrapping}~\cite{bootstrap}. Means and \acp{std} of those fits yield the relevance scores and errors, respectively.

As stated already, our sampling based method is, if at all, just an approximation of the Shapley kernel, but it enables the generation of potentially more visible and realistically occluded pedestrian instances. In the following experiments, we will evaluate whether we can approximate the Kernel\ac{SHAP} results with our sampling approach, and if so, whether our method can approximate the Shapley values with fewer samples than Kernel\ac{SHAP}.

\section{Experiments and Results}

In this section, we will perform a few experiments about comparability between Kernel\ac{SHAP} and our Beta sampling method. Additionally, since the number of samples is the crucial parameter that impacts the evaluation speed of both methods, we observe the stability of the relevance scores under small sample sizes. As a test model, we use a RetinaNet50~\cite{retinanet} object detection model trained on pedestrians from the EuroCity Persons~\cite{eurocitypersons} dataset. Since we could use any model, the training details do not matter here.

\subsection{Local Explanations}

\begin{figure}[tbh]
    \centering
    \includegraphics[width=\textwidth]{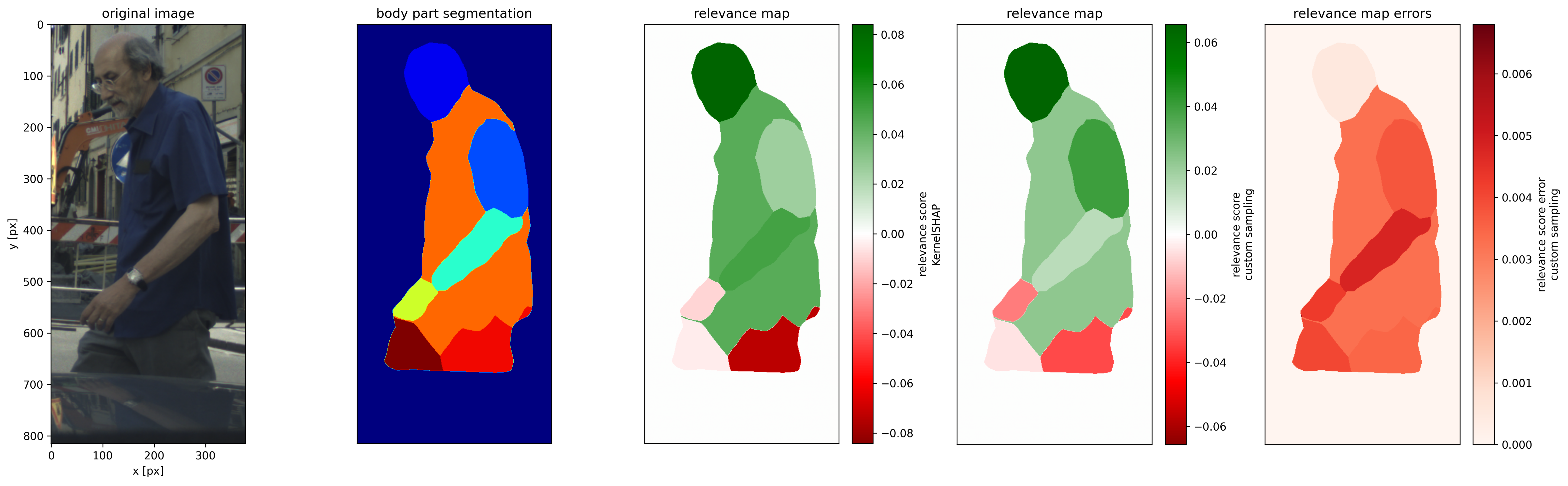}
    \caption{Exemplary body part segmentation by BodyPix~\cite{bodypix_github} and corresponding body part relevance maps of Kernel\ac{SHAP} (middle plot) and our sampling method (second from right). Additionally, an error map for our method is shown in the rightmost plot.}
    \label{fig:relevance_map}
\end{figure}

We evaluate Kernel\ac{SHAP} and our method by using our superpixel model for an example image from the EuroCity Persons dataset. For both methods, \num{2048} samples were drawn. In this case, the superpixel model uses the \textit{inpaint} method to hide the body parts. The original image, segmentation map and the resulting relevance maps are shown in Figure~\ref{fig:relevance_map}. In addition, we show the error map of our Beta sampling method.

We notice that the relevance maps calculated by Kernel\ac{SHAP} and our method are similar. Nevertheless, they show some minor differences. One problem in \ac{XAI} is that there is no \enquote{ground truth} explanation, especially not for model-agnostic methods. Thus, we treat the Kernel\ac{SHAP} results as the standard and try to compare our results with it because Kernel\ac{SHAP} has a heavier game theoretical basement due to the Shapley formalism.

\subsection{\enquote{Global} Explanations}

\begin{figure}[tbh]
    \centering
    \includegraphics[width=0.6\textwidth]{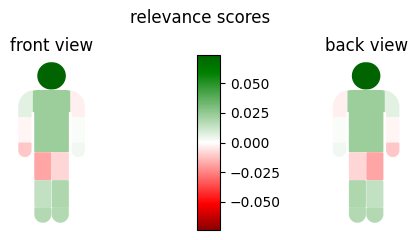}
    \caption{Example of body part relevance maps for \enquote{global} model explanation. Since we have multiple instances now, a human pictogram with color-coded body parts serves as a relevance map.}
    \label{fig:relevance_maps_pictogram}
\end{figure}

Kernel\ac{SHAP} and our method are, per se, local explanation methods. Nevertheless, it could be interesting to investigate, how the model under investigation behaves generally on the majority of (pedestrian) instances. An easy way to do this is to analyze a representative selection of pedestrian instances and average their relevance scores for each body part, where fully occluded body parts are ignored. In our experiments, we take the biggest pedestrians regarding \ac{bbox} area in the dataset of choice. In \ac{AD} street scene datasets, pedestrians are usually quite small, i.e., having a low resolution, so that the segmenting capabilities of BodyPix are even more limited. If high-resolution data is available, also different selection might make sense, e.g., average the biggest, intermediate, and smallest instances separately. To visualize the results intuitively, we color-code the respective body parts by their average relevance scores in a pictogram of a human body. An example is shown in Figure~\ref{fig:relevance_maps_pictogram}. Note that is not a global explanation strategy in the proper sense, which is the reason for the inverted commas in this section's title.

\subsection{Efficient Sampling}

\begin{figure}[H]
    \centering
    \includegraphics[width=\textwidth]{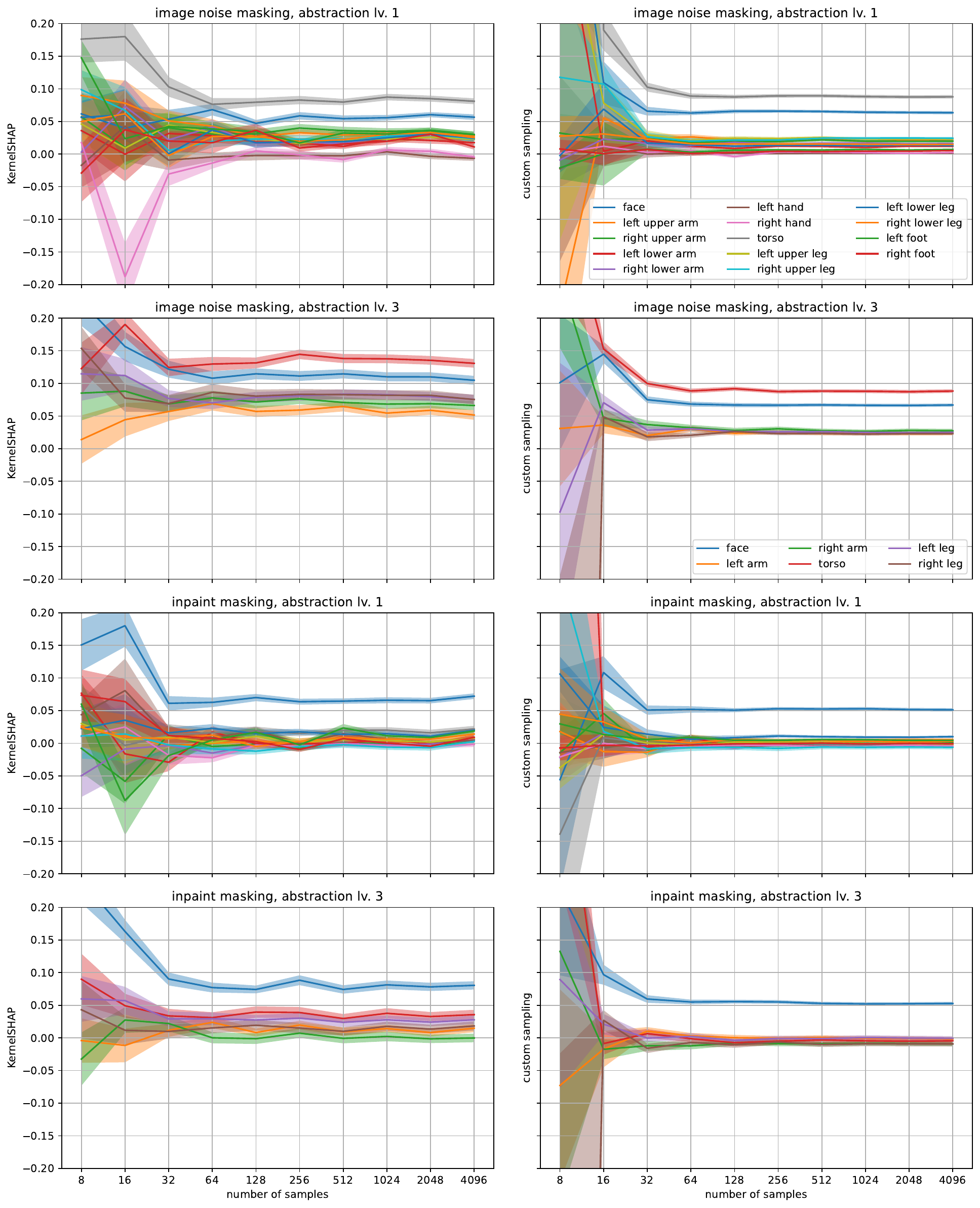}
    \caption{Results of the sampling experiment for abstraction levels 1 and 3, image noise and inpaint masking, and Kernel\ac{SHAP} and our custom Beta sampling method. For each sampling size, the solid lines are the mean relevance scores for the biggest 100 pedestrians in the EuroCity Persons dataset. Transparent bands show the respective \acp{std} of the means.}
    \label{fig:shap_vs_custom}
\end{figure}

In this experiment, we now want to see, how many samples we need at least, to get a fairly stable relevance score determination. We perform these experiments by using the first and third abstraction degree of body parts (see~Figures~\ref{fig:abst_levels:1}~and~\ref{fig:abst_levels:3}) and use inpaint and image noise masking. As sampling sizes, we use powers of \num{2} from \numrange{8}{4096}. In order to also cover, how the methods perform for different pedestrians, we, again, perform the sampling on the \num{100} biggest pedestrians in the EuroCity Persons dataset regarding \ac{bbox} area. Among those, \num{2} could not be segmented properly, so that \num{98} pedestrians contribute in the final results shown in Figure~\ref{fig:shap_vs_custom}. In both abstraction degrees, body parts that do not undergo a merging with other body parts, namely face and torso, have agreeing relevance scores. A remarkable fact is, that the relevance scores differ among the masking methods. For instance, the torso has a significantly higher relevance score for the image noise masking than for the inpaint masking. However, comparing Kernel\ac{SHAP} with our Beta sampling method, we observe that Beta sampling yields more stable results. At \num{64} samples per pedestrian, the Beta sampling already gives results comparable to the higher sampling sizes. Kernel\ac{SHAP}, however, needs more samples to give converging relevance scores, if they converge at all. Conclusively, all experiment show, that our test model mainly focuses on torso and face regions which means that the clear presence or visibility of torso and head mainly drives the pedestrian detection quality.

\section{Discussion}

As already mentioned, Kernel\ac{SHAP} and our Beta sampling method yield comparable relevance scores. This makes sense by looking at the similar sampling properties. The Shapley kernel prefers samples with either very few or very many visible body parts, as shown in~\cite{shap}. Even if the Beta sampling does not follow the Shapley properties exactly, the \ac{pdf} is concentrated on \num{0} and \num{1} and, thus, samples are similar but with the difference of being non-binary. Therefore, we could say that the Beta sampling method is a continuation, or interpolation, of the Shapley kernel sampling.

The experiments show that our Beta sampling method requires fewer samples for robust relevance score assessment than Kernel\ac{SHAP}. Note that the two introduced methods in this work are local explanation methods per se. In order to gain insights into the global explainability of the test model, many local evaluations must be carried out, as we did in the experiments with many pedestrian instances. Thus, our method enables time-efficient analysis for large-scale datasets.

Nevertheless, a shortcoming in this work is the usage of the BodyPix body part segmentation model for the pedestrian detection. BodyPix is mainly used for high-resolution footage of human bodies. However, in street scene data, pedestrians are usually quite far away and, thus, have bad resolutions. Therefore, BodyPix can hardly segment proper body parts for those pedestrians. Additionally, BodyPix cannot discriminate different pedestrian instances, which is problematic in pedestrian detection since pedestrian occur in groups quite often and \acp{bbox} overlap. This is why this work has to be seen as a proof-of-concept for the pedestrian detection use case. It is desirable to use our methods with datasets having available proper body parts and instance segmentation maps. To our knowledge, there is currently no such large-scale street scene dataset available. However, our method could be applied to other tasks concerning (street) scene understanding. The most critical bottleneck is the availability of labels but, in case of uncertainty, one could also stick to fixed image regions like rectangular shaped superpixels. This could be also an approach if the semantic connected between image regions is not as clear, as in the case of, for instance, body parts of pedestrians. 

\section{Conclusion}

Our work demonstrates, that Kernel\ac{SHAP} can be adapted to \ac{OD} use cases. Moreover, the robustness can be increased by using non-binary sampling that is still similar to the Shapley kernel sampling. Our sampling method approximates the Shapley values using fewer samples than Kernel\ac{SHAP} which make evaluation of large-scale object detection on large-scale datasets more efficient. With specific reference to our application of pedestrian detection, it must be noted that BodyPix can only be used to a very limited extent for street scene shots due to the low resolution of the pedestrians. A possible starting point for further research would therefore be the use of simulation data, for which detailed semantic and instance segmentation maps are possibly rather available. Additionally, simulation data can further enrich the analysis by considering attributes beyond body parts like accessories or vehicles like bikes, wheelchairs, buggies, etc. Simulations also enable to gain data tailored to answer specific questions or scenarios that rarely appear in real-world data.

\section{Abbreviations}

\begin{acronym}
    \acro{ABN}{Attention Branch Network}
    \acro{AD}{autonomous driving}
    \acro{AI}{artificial intelligence}
    \acro{bbox}[bbox]{bounding box}
    \acroplural{bbox}[bboxes]{bounding boxes}
    \acro{CAV}{Concept Activation Vector}
    \acro{CNN}{Convolutional Neural Network}
    \acro{Grad-CAM}{Gradient-weighted Class Activation Mapping}
    \acro{GAN}{Generative Adversarial Network}
    \acro{CBR}{Case-Based Reasoning}
    \acro{DICE}{Sørensen-Dice Coefficient}
    \acro{LIME}{Local Interpretable Model-agnostic Explanations}
    \acro{OD}{object detection}
    \acro{pdf}{probability density function}
    \acro{SHAP}{Shapley Additive Explanations}
    \acro{std}{standard deviation}
    \acro{TCAV}{Testing with Concept Activation Vector}
    \acro{XAI}{Explainable Artificial Intelligence}
\end{acronym}
%
%
%
\bibliographystyle{splncs04}
\bibliography{references}

\end{document}